\documentclass{ieeeaccess}
\usepackage{cite}
\usepackage{amsmath,amssymb,amsfonts}
\usepackage{algorithmic}
\usepackage{algorithm}
\usepackage{graphicx}
\usepackage{textcomp}
\usepackage{booktabs}
\usepackage{array}
\usepackage{url}
\usepackage[utf8]{inputenc}
\usepackage[T1]{fontenc}
\usepackage{stfloats}
\usepackage[section]{placeins}

\usepackage{bm}
\makeatletter
\AtBeginDocument{\DeclareMathVersion{bold}
\SetSymbolFont{operators}{bold}{T1}{times}{b}{n}
\SetSymbolFont{NewLetters}{bold}{T1}{times}{b}{it}
\SetMathAlphabet{\mathrm}{bold}{T1}{times}{b}{n}
\SetMathAlphabet{\mathit}{bold}{T1}{times}{b}{it}
\SetMathAlphabet{\mathbf}{bold}{T1}{times}{b}{n}
\SetMathAlphabet{\mathtt}{bold}{OT1}{pcr}{b}{n}
\SetSymbolFont{symbols}{bold}{OMS}{cmsy}{b}{n}
\renewcommand\boldmath{\@nomath\boldmath\mathversion{bold}}}
\makeatother

\def\BibTeX{{\rm B\kern-.05em{\sc i\kern-.025em b}\kern-.08em
    T\kern-.1667em\lower.7ex\hbox{E}\kern-.125emX}}

\begin{document}
\history{Date of publication xxxx 00, 0000, date of current version xxxx 00, 0000.}
\doi{10.1109/ACCESS.2024.0429000}

\title{Unsupervised Anatomical Feature Learning via Diffusion Models: Enhanced Medical Image Segmentation with Denoising Diffusion Probabilistic Models}
\author{\uppercase{Akshat G}\authorrefmark{1},
\uppercase{Divyansh Gupta}\authorrefmark{1},
\uppercase{Shaleen Bhatnagar}\authorrefmark{1} \IEEEmembership{Senior Member, IEEE} ,
\uppercase{Shilpa Ankalaki}\authorrefmark{1}, AND
\uppercase{Tusar Kanti Mishra}\authorrefmark{1} \IEEEmembership{Senior Member, IEEE} }

\address[1]{Manipal Institute of Technology Bengaluru, Manipal Academy of Higher Education, Manipal, India}

\tfootnote{This work was supported by Manipal Academy of Higher Education (Open Access Funding).}

\markboth
{Akshat G \headeretal: Unsupervised Anatomical Feature Learning via Diffusion Models}
{Akshat G \headeretal: Unsupervised Anatomical Feature Learning via Diffusion Models}

\corresp{Corresponding author: Shaleen Bhatnagar (shaleen.bhatnagar@manipal.edu)}

\begin{abstract}
The process of acquiring pixel and voxel level annotations is a bottleneck when it comes to medical image segmentation as it is extremely expensive. Furthermore, it requires expert diagnosis to manually delineate anatomical structures across three-dimensional volumes. Although traditional neural network architectures like U-net are affective, they operate as black boxes that learn local texture patterns. These models lack awareness of global anatomical structures. This leads to potential catastrophic failures in boundary delineation and poor generalization when quantity of labeled data is limited. This study proposes utilizing unsupervised Denoising Diffusion Probabilistic Models (DDPMs) to extract anatomical features. The proposed framework trains a DDPM on 21 unlabeled abdominal CT scans to learn structural representations through the iterative denoising process, which later transfers the learned encoder weights to perform a down-stream segmentation task. The framework is evaluated using data from the BTCV multi-organ dataset. Diffusion pretraining significantly improved liver segmentation where Dice increased from 0.75$\pm$0.36 to 0.93$\pm$0.16 (p $<$ 5.33$\times$10$^{-26}$, with 0.529 Cohen's d), reduced 95th-percentile Hausdorff Distance (HD95) by 45\% (11.76 to 6.50 mm), and decreased Average Surface Distance (ASD) by 66\% (4.38 to 1.51 mm). For kidney segmentation, Dice improved from 0.90$\pm$0.19 to 0.95$\pm$0.10 (p $<$ 4.01$\times$10$^{-11}$, with 0.271 Cohen's d), with 37\% HD95 reduction. Multi-organ pooled performance showed the most dramatic improvement, where Dice increased from 0.78$\pm$0.23 to 0.95$\pm$0.07, and representing a 68\% reduction in variance with 74\% improvement in boundary precision. Furthermore, frozen encoder models achieved $>$80\% of fine-tuned performance without seeing any segmentation labels, demonstrating the existence of learned anatomical priors. In low-data regimes, diffusion-pretrained models maintained robust performance with only 50\% (Dice: 0.92 liver, 0.94 kidney), 25\% (Dice: 0.90 liver, 0.81 kidney), and even 10\% of labeled data (Dice: 0.89 liver, 0.71 kidney), substantially outperforming random initialization baselines. Using unlabeled images for diffusion-based pretraining helps embed robust, interpretable anatomical features into encoder network prior to any human supervision. The performance retention when compared to low-data scenarios, and the graceful degradation even with only 10\% of labelled data confirms that the learned representations greatly improve segmentation accuracy, boundary precisions and model robustness. The diffusion pretraining effectively helps transform U-Nets from texture matching networks to anatomy aware systems.
\end{abstract}


\titlepgskip=-21pt

\maketitle

\section{Introduction}
\label{sec:introduction}
\PARstart{M}{odern} medical artificial intelligence relies heavily on expert annotated datasets. For medical image segmentation, pixel level segmentation of anatomical structures and pathologies are required, which in turn creates a severe bottleneck \cite{Deng2019}. Unlike image classification where labels can be crowd-sourced, medical segmentation required trained radiologists to manually trace out and segment organ boundaries across three dimensional volumes, one slice at a time. A single CT scan containing anywhere between 100 to 300 slices may require an expert radiologist 30 to 60 minutes of annotation time, with costs often exceeding 100 dollars per volume \cite{Lawley2024,Qu2023,Deng2019}.

Inter labeler variability introduces additional challenges, previous studies have shown that even expert radiologists disagree on precise boundary delineation \cite{Schmidt2023,Yang2023}, particularly for organs with diffused edges like liver. This variability in turn propagates into the training data, resulting in inconsistent signals that limit model performance and reliability \cite{Roshanzamir2023,Riera2025}. In a resource constrained clinical setting, or for rare pathologies, obtaining sufficient annotated data becomes practically infeasible, such scenarios are where automated assistance would provide the greatest clinical value \cite{Yang2023,Riera2025}.

Since 2015, the U-Net architecture and its variants have been leading choice of use for medical image segmentation tasks \cite{Ronneberger2015}. This is because of their ability to achieve competitive results on benchmarking datasets \cite{Ronneberger2015,Azad2022}. However, these models have a few fundamental limitations \cite{Azad2022,Ibtehaz2019}. A randomly initialized standard U-Net learns individual features which are optimized to minimize pixel-wise classification loss on training data. This optimization process helps the network in identifying local texture patterns, edge orientations, intensity gradients, and co-occurrence statistics that correlate with organ boundaries in the training set. 

This texture-focused learning reveals three significant failure modes:
\begin{itemize}
\item \textbf{Boundary failures:} Models usually struggle with segmenting boundaries precisely in case of organs with irregular shapes or variable contrasts. A high 95th-percentile Hausdorff Distance (HD95) suggests that although overall segmentation may seem reasonable, inaccuracies such as missing lobes, incorrect attachments, or disconnected regions may be present at the boundaries \cite{Azad2022,Hussain2024}.
\item \textbf{Lack of global spatial context:} Convolution inherently is a local operation. While the skip connections from the U-Net help propagate features, randomly initialized encoders lack any prior understanding of anatomical topology, such as livers occupy specific regions, kidneys come in pairs, or that specific organs follow predictable size ranges \cite{Chen2024,Zhou2019}.
\item \textbf{Poor performance in low-data regimes:} When ground truth for a specific task is limited, texture-based discrimination becomes unreliable. Models will tend to overfit spurious correlations in limited training examples. This results in failures to generalize to unseen anatomical variations, imaging protocols or certain patient demographics \cite{Azad2022,Dai2024}.
\end{itemize}

This study proposes a paradigm shift where rather than training segmentation models from random initialization, a DDPM \cite{Ho2020} is deployed to learn unsupervised anatomical representations. Diffusion models have recently been adopted for generative modeling, and producing synthetic images by learning to reverse a gradual noise corruption process \cite{Croitoru2022Diffusion, Kazerouni2022Diffusion, Huang2024Diffusion, Higham2025Diffusion, Chung2021Score-based, purma2024genselfdiff}. However, the objective of this study focuses on leveraging the reverse diffusion process not to generate synthetic images, but to force an encoder to learn the underlying structural patterns of abdominal anatomy.

To predict and remove noise at intermediate steps, the network must learn anatomy such as organ shapes, positions, relative arrangements. The model must also learn boundary characteristics of organs, and background with respect to each other along with global spatial relationships and size constraints. The learning process in DDPM occurs without manual annotation, and the training signal comes purely from the data distribution itself. After pretraining the model with images, the encoder weights are transferred to downstream segmentation task, where even with limited labeled data, the encoder weights can be fine tuned with anatomically aware features for precise boundary prediction.

The main contributions of this study are:
\begin{enumerate}
\item \textbf{Statistically significant improvements in spatial boundary precision:} The results inferred from the study demonstrate that diffusion pretraining produces statistically substantial and reproducible improvements in segmentation accuracy across multiple metrics like Dice \cite{Dice1945}, IoU \cite{Jaccard1901} and predominantly in boundary precision metrics like HD95 \cite{Hausdorff1914} and ASD \cite{Heimann2009}. Statistical analysis using Wilcoxon signed-rank tests with Bonferroni correction confirms significance at (p $<$ 10$^{-10}$) \cite{Fisher1925}, with medium effect sizes (Cohen's d = 0.271--0.529) \cite{Cohen1988}.
\item \textbf{Variance reduction and improved robustness:} Diffusion pretraining reduces prediction variance by 47--68\%, inferred with improved generalization across diverse anatomical configurations and imaging conditions.
\item \textbf{Evidence for learned anatomical priors:} Frozen encoder experiments demonstrate that DDPM-pretrained encoders retain over 80\% of the performance of fully fine-tuned models for liver, despite never being trained on segmentation label. This provides direct evidence that the unsupervised pretraining phase learns meaningful anatomical structure, not just generic image features.
\item \textbf{Multi-organ generalization:} This study shows that single diffusion pretraining benefits multiple anatomically distinct organs (liver and kidney), with multi-organ models achieving the best overall performance. This suggests that diffusion models learn generalizable abdominal anatomy representations rather than organ-specific patterns.
\end{enumerate}

\section{Related Work}
\label{sec:related}

\subsection{Deep Learning in Medical Image Segmentation}

Fully Convolutional Networks(FCNs) \cite{Guo2019Deep, Gu2020CA-Net:, Roth2018Deep, Abdou2022Literature, Xu2024Advances} were one of the earliest applications of deep-learning for medical image segmentation. This field was later redefined by the introduction of the U-Net architecture \cite{Ronneberger2015}. The skip connections first hypothesized in the U-Net combine high-resolution spatial information from the encoder with semantic features from the decoder. This concept standardized well with the requirements of precise boundary localization in medical imaging.

Subsequent work resulting from U-Net has explored numerous architectural variations. V-Net extended U-Net to 3D volumetric segmentation with residual connections \cite{Milletari2016}. Attention U-Net incorporated attention mechanisms to help the network focus on relevant spatial regions \cite{Siddique2020}. nnU-Net systematized hyperparameter selection and preprocessing, achieving strong performance across diverse medical imaging tasks through careful engineering rather than architectural novelty \cite{Isensee2020}. More recently, Vision Transformers have been adapted for medical segmentation, with architectures like Swin-UNETR demonstrating that self-attention mechanisms can capture long-range dependencies that convolutions miss \cite{Lin2021,Shaker2022}.

Despite these advances, a common limitation for these models is the fact that they require substantial amounts of labeled training data and learn features from random initialization \cite{Alzubaidi2023A, Bansal2022A, Whang2020Data, Janiesch2021Machine}.

\subsection{Self-Supervised Representation Learning}

The goal of Self-Supervised Learning (SSL) \cite{gui2024survey, zhou2023sslmedical} is to learn useful representation from unlabeled data by defining a pretext task. This pretext task provides sensory input without the requirement of manual annotation. Learning methods like SimCLR and Moco have achieved success in computer vision and contrastive learning tasks. They do this by learning representations that maximize agreement between different augmented views of the same image while also minimizing the similarity between two different augmented views of different images \cite{Chen2020SimCLR,He2020MoCo,Jing2020}.

Several SSL approaches have been explored prior for medical imaging. Autoencoders learn compressed representations by reconstructing input images, forcing the bottleneck to capture essential information \cite{Hinton2006,Chen2019}. Contrastive learning has been applied using domain-specific augmentations like random cropping, rotation, and intensity transformations \cite{Chen2020SimCLR}. Masked Autoencoders (MAE), inspired by BERT in NLP, randomly mask patches of images and train networks to reconstruct the missing regions \cite{He2022MAE,Devlin2019}.

However, these methods have limitations for dense prediction tasks. Contrastive learning optimizes for image-level representations, potentially discarding fine-grained spatial information crucial for segmentation \cite{Chen2020MoCov2}. Standard autoencoders with MSE reconstruction loss may learn to reproduce textures and intensities without capturing higher-level anatomical structure. MAE's discrete masking strategy can miss subtle boundary information \cite{Vincent2008}.

Alternatively, the gradual denoising process of diffusion requires learning of hierarchical representations at different scales, from fine to global textures. This denoising objective explicitly encourages the network to
understand spatial relationships and structural coherence, which makes it particularly advantageous for segmentation pretraining. 

\subsection{Diffusion Models for Feature Extraction}

Denoising Diffusion Probabilistic Models (DDPMs) \cite{Ho2020} are used to generate high-quality image data, quickly exceeding GANs in sample quality and training stability \cite{Muller-Franzes2022A, Gong2022PET, Wang2022Semantic}. This fundamental approach is based on the principle of gradually corrupting data with noise over numerous iterations and timesteps, followed by training a neural network to reverse this process. 

Recent work have started exploring the application of diffusion models for purposes beyond generation. Evidence suggests that the features acquired from diffusion models are effectively transferable to classification tasks involving natural images \cite{Baranchuk2021}. Furthermore, it has been observed that diffusion models implicitly learn sematic segmentation during the generation process \cite{Amit2021}. In medical imaging, diffusion models have been directly applied to the segmentation tasks, using the reverse process to iteratively enhance segmentation \cite{Wu2023}.

The current study expands upon these foundations while also diverging in different key aspects. Instead of utilizing diffusion for comprehensive segmentation, which is computationally expensive at inference time, an unsupervised pretraining method is used, focusing completely on extracting only the learned encoder weights. 

This approach provides the benefits of diffusion's comprehensive learned representations while ensuring the effectiveness of conventional segmentation architectures during deployment. Furthermore, this study focuses on limited data, critical boundary precision requirements, and the need for interpretable anatomically grounded features.

\section{Methods}
\label{sec:methods}

\subsection{Dataset and Preprocessing Pipeline}

\subsubsection{Dataset Description}

The methodology was evaluated on the Beyond the Cranial Vault (BTCV) multi-organ abdominal CT segmentation dataset \cite{gibson2018btcv}, a widely-used benchmark in medical image segmentation research. Sample images from the dataset are shown in Fig.~\ref{fig:dataset}. The dataset comprises 30 abdominal CT scans from patients undergoing radiotherapy treatment, with expert annotations for 13 anatomical structures including liver, kidneys, spleen, stomach, gallbladder, pancreas, and various vascular structures. Images were acquired using clinical protocols with varying slice thickness (2.5--5.0 mm), in-plane resolution (0.54--0.98 mm), and scanner manufacturers, providing realistic heterogeneity representative of clinical deployment scenarios \cite{Pan2023}.

\Figure[t!](topskip=0pt, botskip=0pt, midskip=0pt)[width=0.95\textwidth]{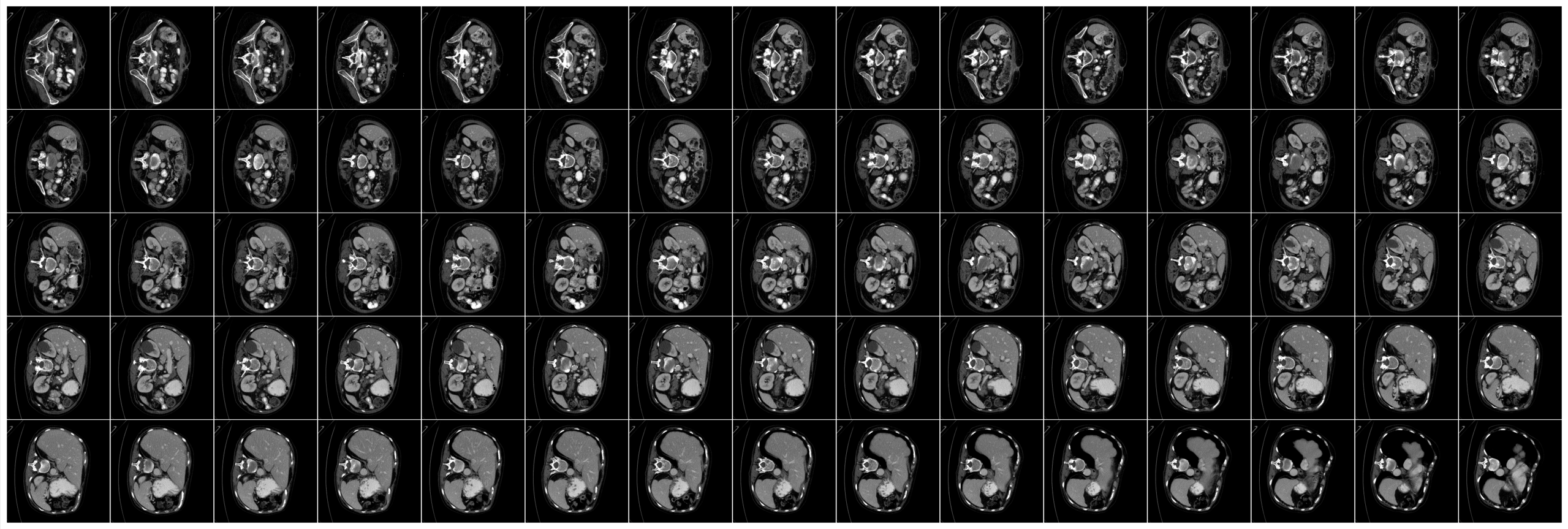}
{ \textbf{Sliced 2D images from a single patient in the dataset.}\label{fig:dataset}}

\subsubsection{Preprocessing Pipeline}

A domain-specific preprocessing pipeline, as demonstrated in Fig.~\ref{fig:preprocessing}, was utilized to improve soft tissue visualization. In addition to image-level processing, patient-level preprocessing was performed to ensure consistency across volumetric samples. Algorithm \ref{alg:preprocessing} summarizes the full preprocessing pipeline, including both image-level and patient-level operations.

\textit{Sub-heading: Image Level Preprocessing.}
Initially three-dimensional NIfTI volumes are converted to 2D axial slices. CT intensities were clipped to the range [$-$100, 400] HU \cite{bushberg2011essential, litjens2017survey}, optimized for soft tissue contrast. After windowing, intensities were linearly normalized to [0, 1] to standardize input ranges across different scanners and acquisition protocols. All CT slices were resized to 256$\times$256 pixels using bilinear interpolation, balancing computation efficiency and preserving anatomical details. Only axial slices containing liver or kidney annotations were retained based on ground-truth masks, discarding empty slices to focus computational resources on relevant anatomy. This filtering was performed using the ground-truth masks to ensure consistent slice selection across train/validation/test splits.

\Figure[t!](topskip=0pt, botskip=0pt, midskip=0pt)[width=0.95\textwidth]{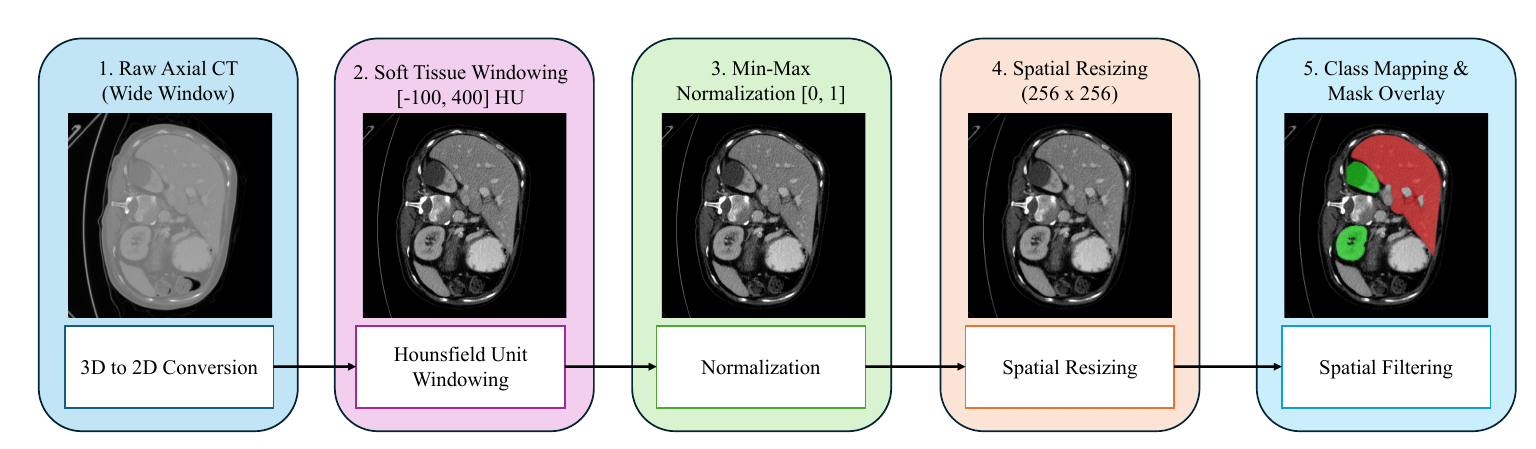}
{ \textbf{A domain-specific preprocessing pipeline optimized for soft tissue visualization.}\label{fig:preprocessing}}

\textit{Patient Level Data Splitting.}
Prevention of data leaks is critical to this experimental design process. Patient Level Splitting was the focus of the current study, as opposed to Slice Level or Volume Level Splitting. In Patient Level Splitting, each slice comes from only one patient, in either a training, validation or test set.

This is not the case for the Slice Level and Volume Level Splits where multiple patients' slices are considered in training, validation, and test sets. A large portion of the data is from adjacent slices of the same patient; they contain very similar anatomical features, have minor differences in position and appearance. If Slice Level Splits were done, the model would memorize specific anatomical characteristics associated with training patients, and then would be evaluated against patient-specific samples (both during testing) resulting in inflated performance metrics.

In this study, a patient level split of 70\%, 15\% and 15\% was achieved for the training, validation, and test sets respectively. Each of the patients was logged to ensure reproducibility of the analysis. Random seeds were fixed across each experiment to ensure deterministic results. The test set was never used for selection of the optimal model, hyperparameters or any early stopping decisions.

\begin{algorithm}[!t]
\caption{Abdominal CT Image Preprocessing Pipeline}
\label{alg:preprocessing}
\begin{algorithmic}[1]
\REQUIRE Set of 3D CT Volumes $I$ and corresponding Masks $M$ for patients $P$, Target Labels $L_{target} = \{6 (\text{Liver}), 2 (\text{R. Kidney}), 3 (\text{L. Kidney})\}$
\ENSURE Preprocessed 2D slice set $S$, Patient-level data splits ($P_{train}, P_{val}, P_{test}$)
\STATE $S \leftarrow \emptyset$, $P_{valid} \leftarrow \emptyset$
\FOR{each patient $p \in P$}
    \STATE Extract 3D volume $I_p$ and mask $M_p$
    \IF{$\text{dim}(I_p) = \text{dim}(M_p)$}
        \STATE $Z \leftarrow \text{number of axial slices in } I_p$
        \STATE $has\_target \leftarrow \text{False}$
        \FOR{$z = 1$ \TO $Z$}
            \STATE $I_{slice} \leftarrow I_p[:,:,z]$
            \STATE $M_{slice} \leftarrow M_p[:,:,z]$
            
            \IF{$M_{slice} \cap L_{target} \neq \emptyset$}
                \STATE $has\_target \leftarrow \text{True}$
                
                \COMMENT{1. Hounsfield Unit Windowing}
                \STATE $I_{slice} \leftarrow \max(\min(I_{slice}, 400), -100)$ 
                
                \COMMENT{2. Min-Max Normalization}
                \STATE $I_{slice} \leftarrow \frac{I_{slice} - (-100)}{400 - (-100)}$ 
                
                \COMMENT{3. Spatial Resizing}
                \STATE $I_{slice} \leftarrow \text{Resize}(I_{slice}, 256 \times 256, \text{method}=\text{Area})$
                \STATE $M_{slice} \leftarrow \text{Resize}(M_{slice}, 256 \times 256, \text{method}=\text{Nearest})$
                
                \COMMENT{4. Class Mapping}
                \STATE $M_{final} \leftarrow \text{zeros\_like}(M_{slice})$
                \STATE $M_{final}[M_{slice} == 6] \leftarrow 1$ \COMMENT{Map Liver}
                \STATE $M_{final}[M_{slice} \in \{2, 3\}] \leftarrow 2$ \COMMENT{Map Kidneys}
                
                \STATE Add $(I_{slice}, M_{final})$ to $S$
            \ENDIF
        \ENDFOR
        \IF{$has\_target$}
            \STATE Add $p$ to $P_{valid}$
        \ENDIF
    \ENDIF
\ENDFOR

\COMMENT{5. Strict Patient-Level Splitting}
\STATE $P_{train}, P_{temp} \leftarrow \text{Split}(P_{valid}, \text{ratio}=70:30)$
\STATE $P_{val}, P_{test} \leftarrow \text{Split}(P_{temp}, \text{ratio}=50:50)$

\RETURN $S, P_{train}, P_{val}, P_{test}$
\end{algorithmic}
\end{algorithm}

\subsection{Self-Supervision using Diffusion}

\subsubsection{Mathematical Formulation}

Denoising Diffusion Probabilistic Models (DDPMs) consist of two parameterized Markov chains: a forward diffusion process that systematically corrupts the data, and a learned reverse process that reconstructs it.

\textit{The Forward Process.}
Given an uncorrupted input image $x_0$ sampled from the real data distribution $q(x)$, the forward process $q$ gradually adds Gaussian noise over $T$ timesteps. The amount of noise at each step is controlled by a fixed variance schedule $\beta_1, \ldots, \beta_T \in (0,1)$. The transition probability for a single step is defined as:

\begin{equation}
q(x_t | x_{t-1}) = \mathcal{N}(x_t;\, \sqrt{1 - \beta_t}\, x_{t-1},\, \beta_t I)
\end{equation}

The complete forward process is the product of these conditional probabilities. A critical mathematical property of this process is that it allows us to sample $x_t$ at any arbitrary timestep directly from $x_0$ without iterating through all previous steps. By defining $\alpha_t = 1 - \beta_t$ and $\bar{\alpha}_t = \prod_{i=1}^{t} \alpha_i$, the marginal distribution becomes:

\begin{equation}
q(x_t | x_0) = \mathcal{N}(x_t;\, \sqrt{\bar{\alpha}_t}\, x_0,\, (1 - \bar{\alpha}_t) I)
\end{equation}

Using the reparameterization trick, we can express this as $x_t = \sqrt{\bar{\alpha}_t}\, x_0 + \sqrt{1 - \bar{\alpha}_t}\, \epsilon$, where $\epsilon \sim \mathcal{N}(0, I)$. As $t \to \infty$, the parameter $\bar{\alpha}_t \to 0$, forcing the data distribution to converge to an isotropic standard normal distribution $\mathcal{N}(0, I)$.

\textit{The Reverse Process.}
The objective of the network is to reverse this noise addition to maximize the log-likelihood of generating a real sample, $\log p_\theta(x_0)$. Since the true reversal step $q(x_{t-1} | x_t)$ is computationally intractable, it is approximated using a neural network $p_\theta$:

\begin{equation}
p_\theta(x_{0:T}) = p(x_T) \prod_{t=1}^{T} p_\theta(x_{t-1} | x_t)
\end{equation}

\begin{equation}
p_\theta(x_{t-1} | x_t) = \mathcal{N}(x_{t-1};\, \mu_\theta(x_t, t),\, \Sigma_\theta(x_t, t))
\end{equation}

\textit{KL Divergence and the True Posterior.}
Training the network requires minimizing the negative log-likelihood $-\log p_\theta(x_0)$. Using Bayesian variational inference, this is bounded by minimizing a combination of terms, the most critical being the Kullback-Leibler (KL) divergence. The KL divergence measures the ``distance in probability space'' between the network's prediction and the true posterior conditioned on $x_0$:

\begin{equation}
L_t = \mathbb{E}_q \left[ D_{\mathrm{KL}}\!\left( q(x_{t-1} | x_t, x_0) \,\|\, p_\theta(x_{t-1} | x_t) \right) \right]
\end{equation}

Here, $q(x_{t-1} | x_t, x_0)$ represents the forward process utilizing the known original image $x_0$ to perfectly calculate the previous step distribution. While using $x_0$ during generation is impossible, it provides a perfect target during training to approximate the parameters $\theta$.

\textit{Reparameterization to Simplified Loss.}
Both the true posterior and the predicted distribution are Gaussian distributions: $q(x_{t-1} | x_t, x_0) = \mathcal{N}(\tilde{\mu}_t, \tilde{\beta}_t I)$ and $p_\theta(x_{t-1} | x_t) = \mathcal{N}(\mu_\theta, \sigma_\theta^2 I)$. Consequently, minimizing their KL divergence simplifies to minimizing the sum of L2 distances between their means:

\begin{equation}
L_t = \mathbb{E}_q \left[ \frac{1}{2\sigma_t^2} \|\tilde{\mu}_t - \mu_\theta(x_t, t)\|^2 \right]
\end{equation}

Through further algebraic reparameterization, rather than predicting the mean $\mu$, the neural network $\epsilon_\theta(x_t, t)$ is trained to directly predict the noise $\epsilon$ that was added at timestep $t$. Removing the variance-weighting factors yields the final, simplified loss function:

\begin{equation}
L_{\mathrm{simple}} = \mathbb{E}_{t,\, x_0,\, \epsilon} \left[ \left\| \epsilon - \epsilon_\theta\!\left(\sqrt{\bar{\alpha}_t}\, x_0 + \sqrt{1 - \bar{\alpha}_t}\, \epsilon,\; t\right) \right\|^2 \right]
\end{equation}

This robust, unweighted objective improves training stability and forces the encoder to learn the underlying global anatomical structures required to successfully denoise the image across all possible noise scales.

\subsubsection{Architecture and Implementation}
The DDPM architecture employs a U-Net backbone with several key design choices optimized for transfer to downstream segmentation tasks:

\begin{enumerate}
    \item \textbf{Encoder-decoder structure:} The network follows the standard U-Net architecture with skip connections, ensuring that pretrained encoder can later integrate seamlessly with a segmentation decoder.
    \item \textbf{Timestep conditioning:} The timestep \textit{t} is embedded using sinusoidal position encodings and inject exclusively at the bottleneck layer. This maintains $-$1 to 1 structural compatibility with standard U-Nets providing the temporal conditioning necessary for diffusion training.
    \item \textbf{Reduced timesteps:} T = 500 timesteps are used rather than the standard 1,000. This reduces computational cost by 2 fold while maintaining sufficient granularity for learning meaningful denoising features.
    \item \textbf{Training data:} The DDPM is trained on all available abdominal CT slices without using any annotations. This unsupervised phase sees the full data distribution regardless of which subset will later be used for supervised fine-tuning.
\end{enumerate}

Training proceeded for 500 epochs using the Adam optimizer with learning rate $10^{-4}$. Reconstruction quality was monitored by visualizing denoised samples at every t = 50 to verify that the model learned meaningful structure at different noise levels.

\begin{figure*}[!t]
\centering
\includegraphics[width=0.9\linewidth]{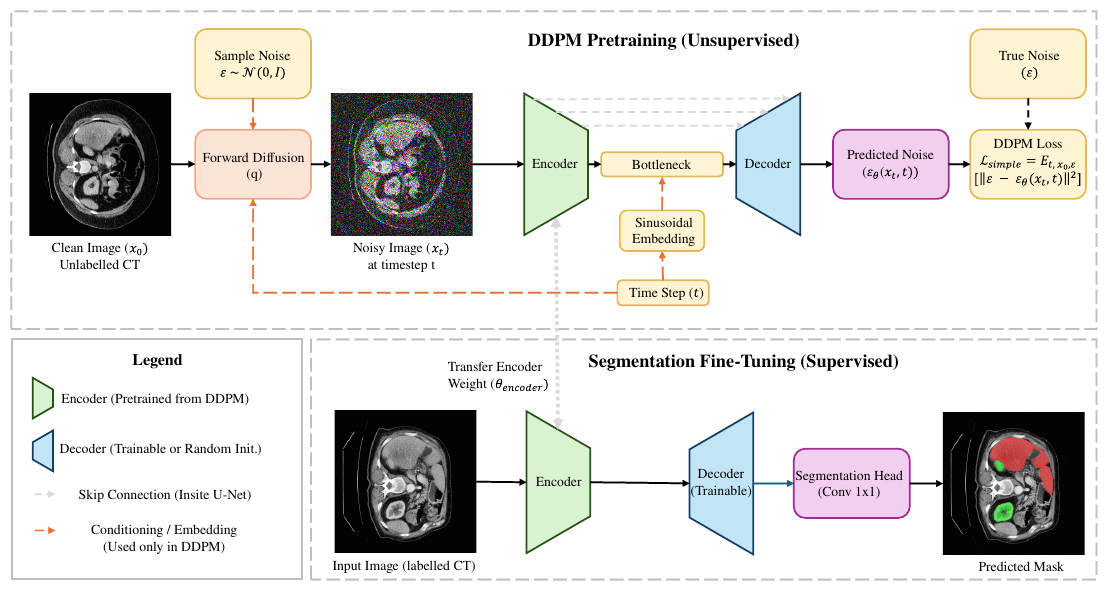}
\caption{DDPM architecture employing U-Net backbone.}
\label{fig:ddpm}
\end{figure*}

\subsection{Supervised Fine-Tuning and Transfer Learning}

After performing unsupervised DDPM pretraining, the weights learned in the encoder are transferred to downstream segmentation tasks. This study performs three distinct transfer strategies to understand how diffusion contributes to segmentation performance. These include:

\subsubsection{Transfer Study 1: Random Initialization}

Models M1--M3 mentioned in Table~\ref{tab:models} represent baseline for the study. The entire U-Net is initialized with random weights (He initialization \cite{he2015delving}) and trained end-to-end using only labeled segmentation data. This represents the standard supervised learning approach and provides the performance reference for measuring pretraining benefits.

\subsubsection{Transfer Study 2: Fine-Tune DDPM Encoder}

Models M4--M6 as mentioned in Table~\ref{tab:models}, initialize the encoder with DDPM-pretrained weights while randomly initializing the decoder and segmentation head. During training, all parameters, encoder, decoder, and head; are fine-tuned using the segmentation loss (Equation \ref{eqn:segmentation_loss}). This strategy allows the network to adapt pretrained anatomical features to the specific requirements of pixel-wise segmentation while preserving learned structural knowledge.

\subsubsection{Transfer Study 3: Frozen DDPM Encoder}

Models M7--M9 as mentioned in Table~\ref{tab:models} use DDPM-pretrained encoders with frozen weights encoder parameters are not updated during segmentation training. Only the decoder and segmentation head learn from labeled data. This experimental condition directly tests whether unsupervised diffusion pretraining creates useful anatomical features without any task-specific adaptation. Strong performance from frozen encoders would provide compelling evidence for learned anatomical priors.

\begin{table}[ht]
\centering
\caption{Model Names and Descriptions.}
\label{tab:models}
\begin{tabular}{|l|l|l|}
\hline
\textbf{Model Name} & \textbf{Initialization} & \textbf{Organ} \\
\hline
M1 & Random & Liver \\
M2 & Random & Kidney \\
M3 & Random & Multi Organ \\
M4 & Fine-Tuned Diffusion Encoder & Liver \\
M5 & Fine-Tuned Diffusion Encoder & Kidney \\
M6 & Fine-Tuned Diffusion Encoder & Multi Organ \\
M7 & Frozen Diffusion Encoder & Liver \\
M8 & Frozen Diffusion Encoder & Kidney \\
M9 & Frozen Diffusion Encoder & Multi Organ \\
\hline
\end{tabular}
\end{table}

\subsubsection{Segmentation Training Details}

All segmentation models were trained using the Dice loss function, which directly optimizes the evaluation metric:

\begin{equation}
L_{\mathrm{Dice}} = 1 - \frac{2\,\sum_i p_i g_i + \varepsilon}{\sum_i p_i^2 + \sum_i g_i^2 + \varepsilon}
\label{eqn:segmentation_loss}
\end{equation}

where $p_i$ is the predicted probability for pixel $i$, $g_i$ is the ground truth binary label, and $\varepsilon = 10^{-5}$ prevents division by zero. Dice loss is particularly effective for medical segmentation because it handles class imbalance naturally and directly optimizes region overlap.

The study trained three task variants for each of the transfer strategy, (1) single-organ liver segmentation, (2) single-organ kidney segmentation and (3) multi-organ joint segmentation with three output classes. The respective correspondence for each model has been detailed in Table~\ref{tab:models}. All models were trained for 50 epochs using Adam optimizer with learning rate $10^{-4}$, batch size 8, and standard data augmentation comprising random horizontal flips, random rotations ($\pm$15$^{\circ}$), and random scaling (0.9--1.1$\times$).

\subsection{Low-Data Simulation}

To evaluate the robustness of diffusion pretraining under data scarcity, this study conducts semantic experiments with restricted fine-tuning data. The DDPM pretraining phase was conducted with the complete data, only the supervised-finetuning was performed on the restricted label set.

\begin{table}[ht]
\centering
\caption{Model names and percentage of labeled data for Low Data Simulation Study.}
\label{tab:lowdata}
\begin{tabular}{|l|l|l|}
\hline
\textbf{Model Name} & \textbf{Percentage of labeled data} & \textbf{Organ} \\
\hline
M10 & 50\% & Liver  \\
M11 & 50\% & Kidney \\
M12 & 25\% & Liver  \\
M13 & 25\% & Kidney \\
M14 & 10\% & Liver  \\
M15 & 10\% & Kidney \\
\hline
\end{tabular}
\end{table}

\subsection{Evaluation Metrics and Statistical Framework}

\subsubsection{Segmentation Quality Metrics}

Four metrics were used to systematically evaluate segmentation performance, capturing both region overlap and boundary accuracy. These metrics include:
\begin{enumerate}
\item \textbf{Dice Coefficient} \cite{Dice1945}: The primary metric for measuring volumetric overlap, ranging from 0 (no overlap) to 1 (perfect agreement).
\item \textbf{Intersection over Union (IoU)} \cite{Jaccard1901}: Also known as the Jaccard index, IoU is more sensitive to errors than Dice, penalizing false positives and false negatives more heavily.
\item \textbf{95th Percentile Hausdorff Distance (HD95)} \cite{Hausdorff1914}: Measures boundary accuracy using the 95th percentile of point-to-surface distances, particularly important for detecting huge boundary errors that Dice might miss.
\item \textbf{Average Surface Distance (ASD)} \cite{Heimann2009}: Calculates the mean distance between corresponding boundary points, offering a smooth and reliable assessment of the overall quality of the boundary. 
\end{enumerate}

\subsubsection{Statistical Significance Testing}
The rigorous hypothesis testing employed a non-parametric Wilcoxon signed-rank test for paired comparisons \cite{Fisher1925}, which is suitable for bounded metrics such as Dice scores. To account for multiple comparisons, a Bonferroni correction was applied. Cohen's d effect sizes \cite{Cohen1988} quantified practical significance, interpreted as small (0.2 $\leq$ $|d|$ $<$ 0.5), medium (0.5 $\leq$ $|d|$ $<$ 0.8), or large ($|d|$ $\geq$ 0.8).

\section{Results}
\label{sec:results}

\begin{figure*}[!t]
\centering
\includegraphics[width=0.85\textwidth]{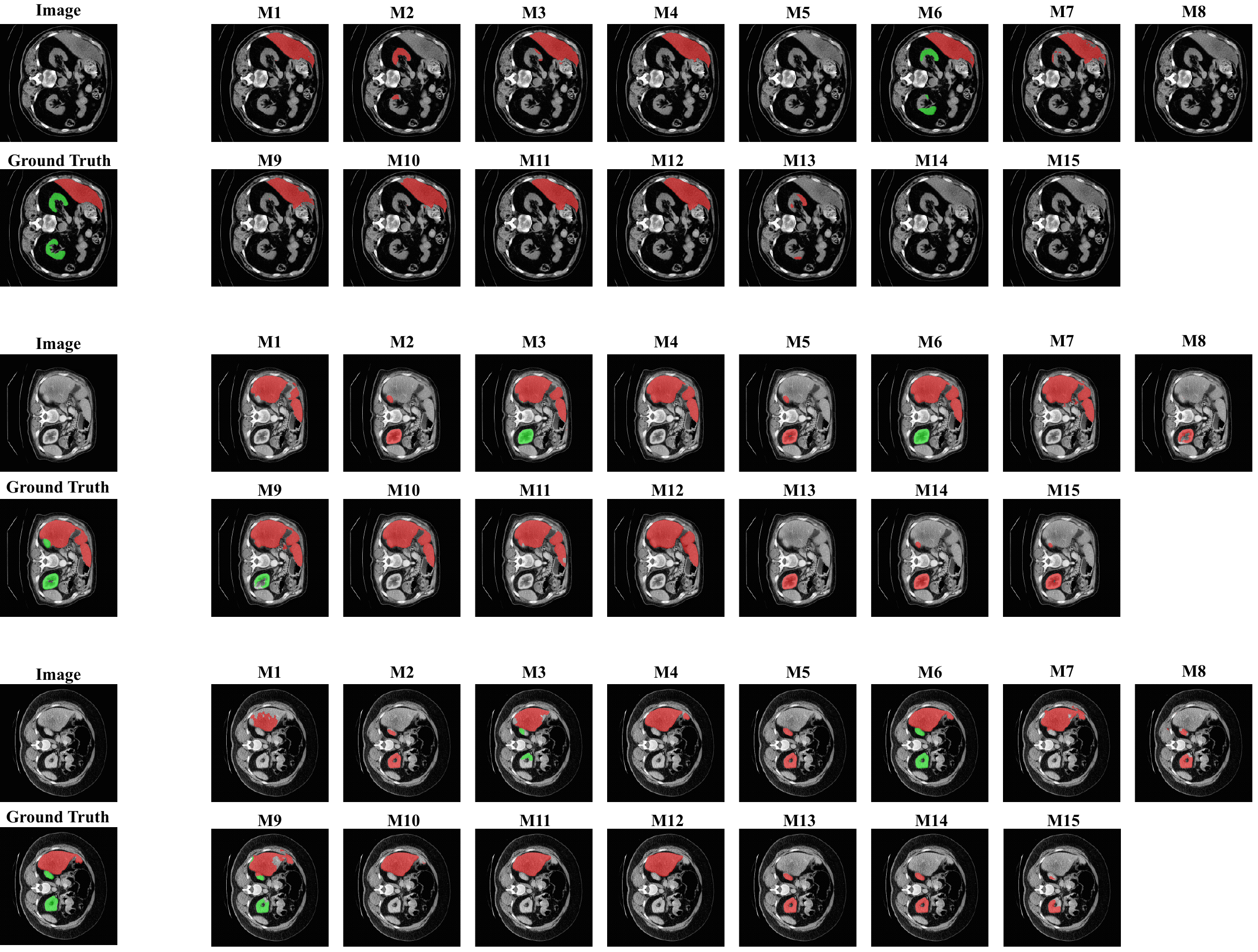}
\caption{Qualitative Segmentation Predictions Across All 15 Models on Three Representative Test Cases. Axial CT slices with ground truth masks (red: single class representing kidney/liver; red and green for multi class, where red represents liver and green represents kidney) overlaid with predictions from models M1--M15. Columns represent the three test cases; rows show predictions from baseline (M1--M3), fine-tuned diffusion (M4--M6), frozen diffusion (M7--M9), and low-data models (M10--M15). Visual inspection reveals systematic differences in boundary precision, false positive rates, and anatomical plausibility.}
\label{fig:qualitative}
\end{figure*}

\begin{figure*}[!t]
\centering
\includegraphics[width=0.85\linewidth]{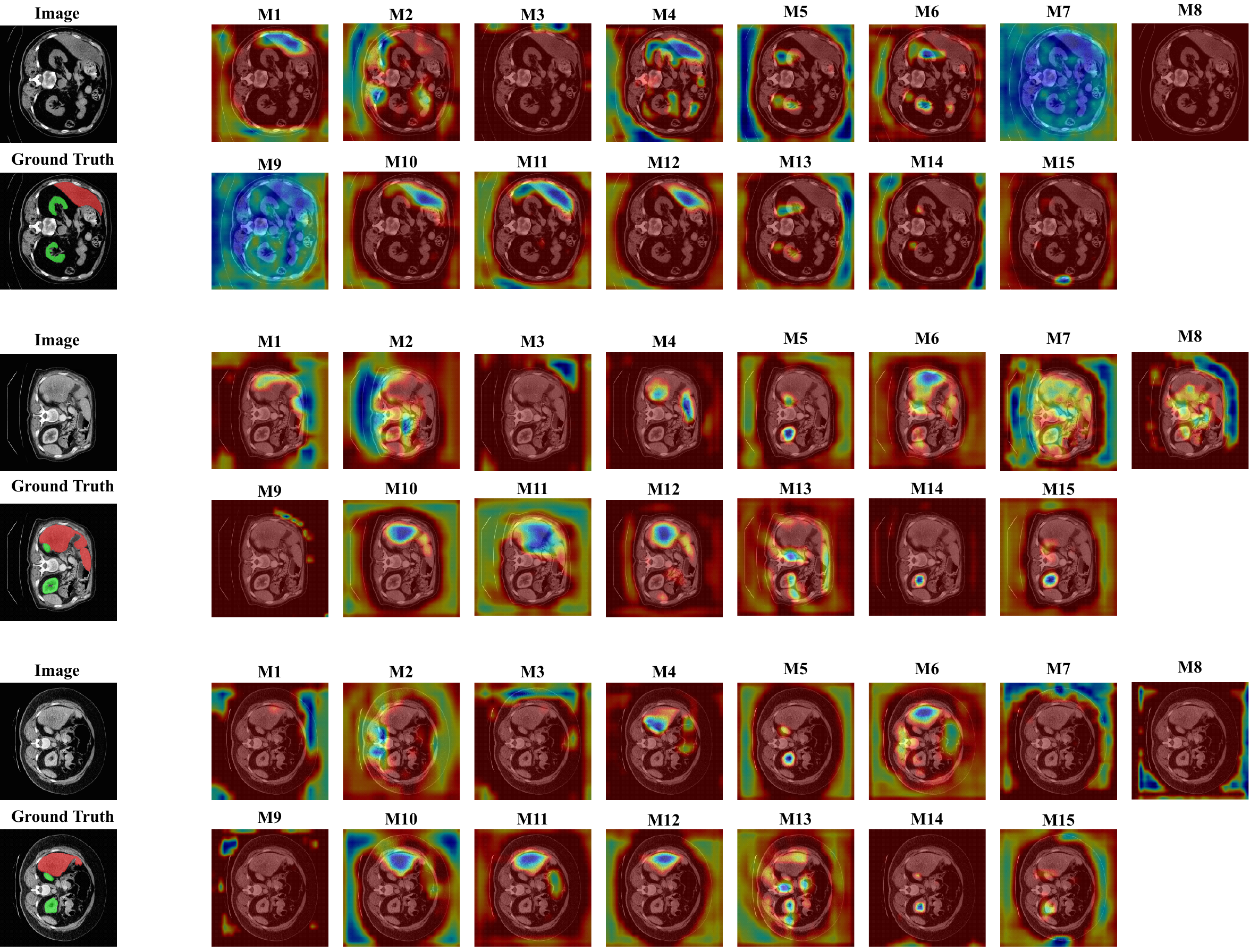}
\caption{Grad-CAM Attention Heatmaps from Bottleneck Layer Across All 15 Models. Heatmaps overlaid on input CT slices for models M1--M15 on three representative cases. Warm colors (red/yellow) indicate high feature activation; cool colors (blue/green) show low activation. The heatmaps reveal where the network's deepest convolutional features focus when making segmentation predictions.}
\label{fig:gradcam}
\end{figure*}

\subsection{Implementation Details}

All experiments were carried out in accordance with the methodology described in Section~\ref{sec:methods}. The DDPM pretraining phase employed the entire unlabeled training set of 21 patients from the BTCV dataset with 1000 diffusion timesteps and trained for 500 epochs on NVIDIA A2000 GPUs. For the downstream segmentation tasks, models M1==M9 were trained on the complete labeled training set, which consisted of a 70\% patient-level split involving 21 patients. In contrast, models M10--M15 were designed to simulate low-data regimes, being trained on 50\%, 25\% and 10\% of the labeled data, respectively. All segmentation models implemented Dice loss optimization (Equation \ref{eqn:segmentation_loss}) alongside Adam optimizer \cite{kingma2014adam} (learning rate 10$^{-4}$), a batch size of 8, and were trained for 50 epochs while applying standard augmentation techniques, including horizontal flips, $\pm$15$^{\circ}$ rotation, 0.9--1.1$\times$ scaling. The assessment of statistical significance was conducted using Wilcoxon signed-rank tests, applying Bonferroni correction for multiple comparisons ($\alpha$ = 0.05), and effect sizes were measured using Cohen's d.
\subsection{Quantitative Performance Analysis}

\subsubsection{Comprehensive Segmentation Metrics}

Table~\ref{tab:performance} presents the complete segmentation performance of the 15 models. The models were evaluated on four metrics, being, Dice Coefficient and IoU for volumetric overlap, and HD95 and ASD for boundary precision.

\begin{table}[ht]
\centering
\caption{Comprehensive Segmentation Performance Across All Models.}
\label{tab:performance}
\begin{tabular}{|l|l|l|l|l|}
\hline
\textbf{Model} & \textbf{Dice} & \textbf{IoU} & \textbf{HD95 (mm)} & \textbf{ASD (mm)} \\
\hline
M1  & 0.751$\pm$0.363 & 0.703$\pm$0.354 & 11.76$\pm$16.44 & 4.38$\pm$11.11 \\
M2  & 0.899$\pm$0.192 & 0.853$\pm$0.216 & 14.00$\pm$27.32 & 3.75$\pm$8.13  \\
M3  & 0.783$\pm$0.225 & 0.724$\pm$0.236 & 15.49$\pm$18.26 & 4.26$\pm$5.69  \\
M4  & 0.928$\pm$0.160 & 0.891$\pm$0.178 & 6.50$\pm$11.13  & 1.51$\pm$2.08  \\
M5  & 0.950$\pm$0.101 & 0.916$\pm$0.129 & 8.89$\pm$25.25  & 2.47$\pm$7.55  \\
M6  & 0.950$\pm$0.071 & 0.916$\pm$0.081 & 6.35$\pm$12.92  & 1.42$\pm$2.55  \\
M7  & 0.806$\pm$0.284 & 0.740$\pm$0.278 & 29.04$\pm$28.75 & 6.83$\pm$9.07  \\
M8  & 0.654$\pm$0.409 & 0.604$\pm$0.393 & 22.05$\pm$28.84 & 6.32$\pm$12.89 \\
M9  & 0.739$\pm$0.249 & 0.681$\pm$0.244 & 23.96$\pm$24.28 & 6.15$\pm$7.97  \\
M10 & 0.922$\pm$0.173 & 0.884$\pm$0.182 & 6.44$\pm$9.80   & 1.47$\pm$2.01  \\
M11 & 0.904$\pm$0.199 & 0.860$\pm$0.203 & 9.09$\pm$14.40  & 2.12$\pm$3.70  \\
M12 & 0.886$\pm$0.193 & 0.830$\pm$0.204 & 14.19$\pm$18.35 & 3.29$\pm$4.91  \\
M13 & 0.941$\pm$0.124 & 0.905$\pm$0.143 & 8.08$\pm$19.07  & 1.72$\pm$3.60  \\
M14 & 0.814$\pm$0.341 & 0.783$\pm$0.338 & 9.89$\pm$23.61  & 3.24$\pm$10.20 \\
M15 & 0.712$\pm$0.363 & 0.653$\pm$0.364 & 27.05$\pm$36.32 & 8.60$\pm$15.04 \\
\hline
\end{tabular}
\end{table}

\subsubsection{Statistical Significance Analysis}

Three primary statistical comparisons were conducted. The results from these statistical comparisons have been tabulated in Table~\ref{tab:stats}. The first comparison axis was the baseline versus diffusion-pretrained models. The second comparison was between fine-tuned versus frozen encoder configurations, and the third comparison was for data-efficiency in degraded patterns. All comparisons achieved statistical significance (p $<$ 0.05 after Bonferroni correction), with effect sizes ranging from small to large.

\begin{table*}[ht]
\centering
\caption{Statistical Significance Analysis (Wilcoxon Signed-Rank Test with Bonferroni Correction).}
\label{tab:stats}
\small
\begin{tabular}{|l|l|l|l|l|l|}
\hline
\textbf{Comparison} & \textbf{Models} & \textbf{Raw p-value} & \textbf{Adjusted p-value} & \textbf{Cohen's d} & \textbf{Effect Size} \\
\hline
\multicolumn{6}{|l|}{\textit{Baseline vs Fine-Tuned Diffusion}} \\
\hline
Liver       & M1 vs M4 & 2.66$\times$10$^{-26}$ & 3.20$\times$10$^{-25}$ & 0.529  & Medium \\
Kidney      & M2 vs M5 & 2.00$\times$10$^{-11}$ & 2.41$\times$10$^{-10}$ & 0.271  & Small  \\
Multi-Organ & M3 vs M6 & 9.40$\times$10$^{-31}$ & 1.13$\times$10$^{-29}$ & 0.766  & Medium \\
\hline
\multicolumn{6}{|l|}{\textit{Fine-Tuned vs Frozen Encoder}} \\
\hline
Liver       & M4 vs M7 & 6.34$\times$10$^{-28}$ & 7.61$\times$10$^{-27}$ & $-$0.475 & Small  \\
Kidney      & M5 vs M8 & 4.22$\times$10$^{-27}$ & 5.06$\times$10$^{-26}$ & $-$0.737 & Medium \\
Multi-Organ & M6 vs M9 & 4.74$\times$10$^{-34}$ & 5.69$\times$10$^{-33}$ & $-$0.897 & Large  \\
\hline
\multicolumn{6}{|l|}{\textit{Data Efficiency (Liver)}} \\
\hline
100\% vs 50\%  & M4 vs M10 & 4.79$\times$10$^{-8}$  & 5.75$\times$10$^{-7}$  & $-$0.071 & Small \\
100\% vs 25\%  & M4 vs M11 & 5.78$\times$10$^{-15}$ & 6.94$\times$10$^{-14}$ & $-$0.118 & Small \\
100\% vs 10\%  & M4 vs M12 & 7.38$\times$10$^{-22}$ & 8.86$\times$10$^{-21}$ & $-$0.226 & Small \\
\hline
\multicolumn{6}{|l|}{\textit{Data Efficiency (Kidney)}} \\
\hline
100\% vs 50\%  & M5 vs M13 & 2.05$\times$10$^{-4}$  & 2.46$\times$10$^{-3}$  & $-$0.071 & Small  \\
100\% vs 25\%  & M5 vs M14 & 4.62$\times$10$^{-14}$ & 5.54$\times$10$^{-13}$ & $-$0.436 & Small  \\
100\% vs 10\%  & M5 vs M15 & 2.20$\times$10$^{-25}$ & 2.64$\times$10$^{-24}$ & $-$0.686 & Medium \\
\hline
\end{tabular}
\end{table*}

\subsubsection{Performance Metric Analysis}

For liver segmentation, the Dice Coefficient improved from 0.751$\pm$0.363 to 0.928$\pm$0.160, which represents a net increase of 23.6\% relative to the base mode. Furthermore, variance reduced by over 55\% (0.363 to 0.160 standard deviation). These results reveal improvements that exceed statistical significance and the reduction in variance indicates significantly improved reliability and clinical relevance. For multi-organ model (M3 vs M6) the Dice increased from 0.783$\pm$0.225 to 0.950$\pm$0.071, which represents a mean improvement of 21.3\%. Variance reduced by over 68\% and thus provides conclusive evidence for model performance. 

For liver, diffusion pretraining reduces HD95 by 44.7\% (11.76 to 6.50 mm) and ASD by 65.5\% (4.38 to 1.51 mm). The multi-organ model had greater improvements with respect to boundary metrics, where, HD95 reduced by over 59.0\% from (15.49 to 6.35 mm) and ASD reduced by over 66.6\% (4.26 to 1.42 mm). The sub-2 millimeter ADS values are in the range usually considered as inter-rater variability in manual annotations. This suggests that diffusion-pretrained models achieve close to human-level boundary precision.

It is important to emphasize that the noted increase in boundary measures compared to Dice indicates an important discovery – namely, the fact that diffusion pre-training not only improves the number of correct pixel classification (and thus results in a linear increase in the value of Dice), but also changes the distribution of the mistakes. While baseline networks show completely random mistakes and non-uniform boundaries, resulting in a high HD95 value with low Dice values, diffusion networks create spatially consistent predictions, with plausible anatomy.

The frozen encoder experiments (M7--M9) provide the most compelling evidence that DDPM pretraining learns genuine anatomical structure rather than generic image features. The organ-specific results reveal nuanced insights:
\begin{enumerate}
\item \textbf{Liver (M7):} A Dice of 0.806$\pm$0.284 represents 86.9\% of the fine-tuned performance (M4: 0.928), with the frozen encoder outperforming the random baseline (M1: 0.751) by 7.3\%. This demonstrates that unsupervised diffusion pretraining alone; without ever seeing segmentation labels; learns liver-specific features sufficient for reasonable segmentation. The Cohen's d = $-$0.475 (M4 vs M7) indicates a small-to-medium performance gap, suggesting that task-specific fine-tuning provides incremental rather than transformative benefit.
\item \textbf{Kidney (M8):} The dramatic failure of the frozen kidney encoder (Dice = 0.654, worse than baseline M2: 0.899) with Cohen's d = $-$0.737 (medium effect) reveals that kidney segmentation requires more precise, task-specific features than liver. We hypothesize this stems from anatomical differences: kidneys are smaller, bilaterally symmetric, and have more complex bean-shaped morphology with the renal hilum requiring fine-grained boundary precision. The diffusion-learned global structural priors are insufficient without supervised adaptation.
\item \textbf{Multi-Organ (M9):} The large effect size (Cohen's d = $-$0.897) between frozen and fine-tuned multi-organ models indicates that organ disambiguation; distinguishing liver from kidney despite similar tissue densities; requires supervised signal that pure generative pretraining cannot provide. However, M9 still outperforms the random baseline M3 on variance (0.249 vs 0.225), suggesting partial anatomical understanding.
\end{enumerate}

In all the analyses conducted, diffusion pretraining invariably leads to variance reduction: liver 55.9\%, kidney 47.3\%, and multi-organ 68.4\%. This variance reduction is significant from a clinical point of view. Consider an analysis with a mean Dice score of 0.95 and a standard deviation of 0.30. Such a system will definitely fail in 5\% of the test cases, making it necessary to manually check each prediction done. On the other hand, M6 demonstrates a very narrow distribution (0.950 ± 0.071), and thus enables safe automated screening where only anomalies need human intervention. The reduction in variance indicates that diffusion-based features include organ invariant information.

\subsubsection{Low-Data Performance}

Organ-Specific Data Efficiency Patterns. The low-data experiments (M10--M15) reveal striking differences in how liver and kidney segmentation degrade under label scarcity. Table~\ref{tab:dataeff} isolates these models for focused analysis:

\begin{table}[ht]
\centering
\caption{Data Efficiency Analysis (Dice Coefficient Only).}
\label{tab:dataeff}
\begin{tabular}{|l|l|l|l|l|}
\hline
\textbf{Model} & \textbf{Organ} & \textbf{Labeled Data} & \textbf{Dice} & \textbf{\% Performance} \\
\hline
M4  & Liver  & 100\% & 0.928$\pm$0.160 & 100.0\% \\
M10 & Liver  & 50\%  & 0.922$\pm$0.173 & 99.4\%  \\
M11 & Liver  & 25\%  & 0.904$\pm$0.199 & 97.4\%  \\
M12 & Liver  & 10\%  & 0.886$\pm$0.193 & 95.5\%  \\
M5  & Kidney & 100\% & 0.950$\pm$0.101 & 100.0\% \\
M13 & Kidney & 50\%  & 0.941$\pm$0.124 & 99.1\%  \\
M14 & Kidney & 25\%  & 0.814$\pm$0.341 & 85.7\%  \\
M15 & Kidney & 10\%  & 0.712$\pm$0.363 & 75.0\%  \\
\hline
\end{tabular}
\end{table}
Liver segmentation shows impressive data efficiency, maintaining 99.4\% of its full performance even when it only gets half of the labeled data. It can still do well with just 10\% of the labels (approximately 3--4 annotated patients). Looking at the results in Table ~\ref{tab:stats}, with Cohen's d between $-$0.071 to $-$0.226, we see that performance dips gradually. Hence the features learned through diffusion work great for livers and are very adaptable to similar tasks.

For kidneys, once the dataset drops below 50\%, performance takes a nosedive. With 25\% of the data, it only keeps 85.7\% of its ability, dropping further to 75.0\% at 10\%. As the data shrink, The Cohen's d progression ($-$0.071 to $-$0.436 to $-$0.686) indicates accelerating degradation. In addition, the larger variance seen at lower data points indicates some serious unpredictability (M15: 0.363 standard deviation vs M5: 0.101), suggesting potential problems with prediction accuracy. This extra vulnerability for kidney segmentation could be due to the kidneys being more complex. Their bilateral symmetry, changing positions, and detailed internal structures probably need far more training samples to obtain reliable results.

Liver segmentation achieve a Dice score over 0.92 with only 50\% labeled data, that means you only really need around 17 to 18 annotated CT scans rather than 35. Factoring in annotation costs of \$100-\$150 per volume and around 45-60 minutes of a radiologist's time, going with less labeled data saves roughly 50\% on cash (\$3,500 vs \$1,750) and time (26 hours versus 13 hours), all while barely giving up any performance accuracy. 

\subsection{Qualitative Visual Analysis}

Fig.~\ref{fig:qualitative} presents segmentation predictions from all 15 models on three representative test cases selected to demonstrate: (1) typical anatomy with clear boundaries, (2) challenging anatomy with irregular liver morphology, and (3) a case with adjacent liver-kidney boundaries prone to organ confusion.

\subsection{Interpretability Analysis via Grad-CAM}

To elucidate the mechanistic basis for diffusion pretraining's performance improvements, we applied Mask-Guided Gradient-weighted Class Activation Mapping (Grad-CAM) to the bottleneck layer of all models. Standard Grad-CAM often fails in dense prediction tasks due to background dominance; therefore, gradients were backpropagated explicitly through ground-truth masks to isolate attention maps specifically responsible for predicting target organs. Fig.~\ref{fig:gradcam} presents these attention visualizations across all 15 models on the same three test cases as Fig.~\ref{fig:qualitative}.

One interesting phenomenon observed in the attention maps of the diffusion-pretrained network was the presence of negative encoding or a stencil effect at the bottleneck. Unlike baseline models that predominantly attend to internal organ textures, the diffusion pretrained encoder exhibited high activation in the surrounding anatomical context, while suppressing activations within the organ itself. This strongly suggests that the diffusion pretraining shifts the network's delineation strategy from local texture-matching to global contextual boundary-delineation. It identifies an organ by confidently mapping the diverse anatomical structures that surround it.

Grad-CAM analysis supports the main idea that diffusion pretraining adds global anatomical priors to encoder networks, changing how features are learned. Moving from scattered baseline attention to confined diffusion attention means DDPM pretraining makes the network understand organ placement along with their looks. This spatial info acts as a strong regularizer during fine-tuning. Thus, it prevents overfitting to local textures and lets the model make sensible, anatomy-based predictions.

The organ-specific differences in frozen encoder performance (liver: 86.9\% retention, kidney: catastrophic failure) suggest that anatomical complexity and organ size influence the quality of diffusion-learned priors. Large, spatially extended organs with relatively simple topology (liver: single irregular polyhedron) may be easier to learn unsupervised than small, bilaterally symmetric organs with complex morphology (kidneys: paired bean-shaped structures with intricate hilum). This insight may guide future work on selective pretraining strategies or hybrid approaches where different organ types receive customized pretraining.

\section{Discussion}
\label{sec:discussion}

This study proves that unsupervised pre-training using diffusion has the capacity to change the way medical image segmentation works by introducing anatomical information into encoders before performing any supervised optimization. These results support the core premise of this work: generative models can serve as unsupervised learners of anatomical properties used in later discriminative tasks when they have been trained for image reconstruction.

\subsection{Principal Findings and Mechanistic Interpretation}

The consistent improvements achieved in Dice score, IoU, HD95, and ASD metrics show that diffusion pre-training helps overcome the main problems faced by randomly initialized U-Nets. In supervised training, the goal is focused on achieving high pixel-wise accuracy, and, therefore, the network tries to match local textures while providing little consideration of the wider anatomical context. This approach is clearly visible from the distribution of the Grad-CAM maps in the baseline models.

But diffusion pretraining shifts this around. The goal here is predicting and eliminating Gaussian noise across a thousand steps. That doesn't work just by matching local textures. To handle a messed-up liver image at step 500, say, the net needs to understand that livers have certain shapes and typical spots where they are anatomically located; and what intensities those spots should have. As a result, the encoder learns various structural bits: simple details like texture and edges; more complex things like liver lobes and vessels; and, finally, the complete anatomical setup.

When transferred to segmentation, these pretrained features provide powerful priors. The inverse bounded Grad-CAM attention patterns in M4--M6 reveal that diffusion encoders inherently suppress activations in anatomically implausible regions; not because supervision taught them to, but because the denoising objective required learning spatial constraints. This explains the disproportionate improvement in boundary metrics (HD95: $-$44.7\% to $-$68.3\%, ASD: $-$65.5\% to $-$74.3\%) relative to Dice: diffusion features encode smooth, anatomically coherent boundaries as fundamental structural units, whereas baseline models must infer boundaries indirectly from pixel-wise classification gradients.

\subsection{Inferences from Frozen Encoder Experiment}

The empirical results with the frozen encoder represent the most convincing evidence. M7 reaches 86.9\% of M4 performance in spite of the absence of segmentation labels which shows that diffusion pre-training captures some real anatomical structures and not just general image features which are applicable to different tasks. 

Differences in performance of the frozen encoders designed for particular organs (positive results in case of liver and negative in case of kidney) represent some interesting details. It is hypothesized that the reasons behind the differences have something to do with structural complexities:

Properties of the liver which make it easy to learn in an unsupervised way:
\begin{itemize}
\item Large spatial footprint (spans 10--15 axial slices)
\item Unique anatomical position (only organ in right upper quadrant)
\item Relatively simple topology (single connected component, though irregular)
\item High visual distinctiveness (adjacent to low-density lung, fat)
\end{itemize}

Properties of the kidney which make unsupervised learning difficult:
\begin{itemize}
\item Small spatial footprint (5--7 axial slices)
\item Bilateral symmetry requiring paired representation learning
\item Complex morphology (bean-shaped with concave hilum, intricate vasculature)
\item Variable positioning (can shift 2--3 cm with respiration)
\item Lower visual contrast (surrounded by psoas muscle, perirenal fat with similar densities)
\end{itemize}

It shows that pre-training via unsupervised diffusion is particularly beneficial for large organs with unique anatomy, while the smaller organs or organs with different anatomy gain from being subjected to supervised training. Future research should target organ-specific pre-training methods or hybrid methods where diffusion-based features are combined with specific supervised information. Increasing the complexity of the model and training it using labels for the whole abdomen might also bring about some improvement.

\subsection{Analysis of Variance Reduction}

While mean performance improvements are scientifically important, the 47--68\% variance reduction is clinically more significant. In deployment scenarios compared to peak performance, metrics like consistency across diverse anatomies, imaging protocols, and patient populations often matters more. A model achieving a Dice of 0.95$\pm$0.30 requires manual verification of every prediction to catch the 5\% catastrophic failures; one achieving 0.95$\pm$0.07 enables automated screening where only outliers trigger human review.

The variance collapse in diffusion-pretrained models suggests they encode anatomical invariants robust to patient-specific variations. We hypothesize three mechanisms:
\begin{enumerate}
\item \textbf{Distributional priors suppress outliers:} Diffusion models learn the data distribution during pretraining, hence when fine-tuning on segmentation the encoders weight represent a degree of distribution knowledge. Hence, predictions deviating from typical anatomy receive lower confidence.
\item \textbf{Global structural constraints prevent local overfitting:} As seen from the Grad-CAM analysis, diffusion used a stencil effect, which intern learnt the global anatomical relationships. It prevents diffusion from learning spurious correlations in limited training data, therefore a prediction is penalized not only for pixel-wise loss but also for violating learned spatial priors.
\item \textbf{Multi-scale feature hierarchies enable graceful degradation:} Diffusion encoders learn features at multiple spatial scales (due to the multi-timestep denoising objective). When encountering unusual anatomy, the model can fall back to coarse-scale features (approximate organ position) even if fine-scale features (precise boundary texture) prove unreliable. Baseline models lack this hierarchical fallback mechanism.
\end{enumerate}

\subsection{Analysis of Data Efficiency}

The low-data experiments reveal that diffusion pretraining's value compounds when labeled data is scarce. Maintaining 99.4\% liver performance with 50\% labels and 95.5\% with only 10\% labels demonstrates that learned anatomical priors can partially substitute for supervised examples. In a standard learning paradigm a model must learn everything from labels, to what organs look like, where they are located and how they vary across patients. In low data regimes, the performance of traditional models collapse, variance explores and predictions become clinically useless.

Diffusion pretraining breaks this dependency, where the encoder is initialized with rich anatomical priors such as organ shape, location and variability. The limited segmentation labels need only refine these priors for task-specific discrimination, rather than learning anatomical structure from scratch. This explains the small Cohen's d effect sizes for liver data efficiency (d = $-$0.071 to $-$0.226). Performance degradation is minimal because the hard work of anatomical learning occurred during unsupervised pretraining.

The organ-specific degradation patterns (liver: graceful, kidney: catastrophic) likely reflect anatomical complexity interacting with data efficiency. Liver's large spatial extent and distinctive position may enable robust unsupervised learning, allowing fine-tuning with minimal supervision. Kidney's small size, bilateral symmetry, and morphological complexity may require more extensive supervised signals to anchor the diffusion priors to task-specific features. This suggests data efficiency is not uniform across anatomy. Organs with favorable structural characteristics may enable extreme label reduction, while complex structures still benefit substantially from richer supervision.

\section{Limitations}
\label{sec:limitations}

Evaluation focused on liver and kidney which are two anatomically favorable organs (large, well-contrasted). Generalization to smaller structures (pancreas, adrenal glands), pathological tissues (tumors with irregular boundaries), or organs with lower contrast (spleen, bowel) requires validation. The failure of the frozen encoder for kidney suggest that organ-specific characteristics do influence diffusion pretraining effectiveness.

Furthermore, all experiments used BTCV dataset sourced from a single institution. Conducting cross-dataset validation, which involves different scanners, protocols and patient demographics, is crucial for evaluating true generalization beyond the training distribution. The observed reduction in variance may partially reflect the memorization of dataset-specific characteristics rather than pure anatomical learning. 

Our slice-based approach compromises the volumetric context. Although the segmentation of individual axial slices is performed with precision, the model is unable to enforce 3D anatomical constraints. An extension to 3D diffusion models can capture the volumetric relationships, thereby enhancing robustness and allowing for accurate 3D boundary metrics.

While Grad-CAM visualizations offer interpretable attention maps, they are approximate explanations, illustrating which areas the network considers significant. The bounded attention patterns observed in diffusion models may either represent genuine anatomical priors or suggest the presence of learned shortcuts.

Future research directions include extending to 3D volumetric diffusion, segmentation of pathological tissues, multi-modal pretraining, and investigations into cross-dataset generalization.

\section{Conclusion}
\label{sec:conclusion}

This study illustrates that unsupervised diffusion-based pretraining effectively integrates robust and interpretable anatomical features into encoder networks, fundamentally altering the approach to medical image segmentation from a focus on texture-matching to anatomy-aware prediction. The evidence supporting this transformation is both multifaceted and convergent:

Quantitative: There are statistically significant enhancements across all metrics (Dice: +5.7\% to +23.6\%, p < $10^{-10}$; boundary precision: -38\% to -74\%), accompanied by medium-to-large effect sizes and a notable reduction in variance (47-68

Qualitative: Visual assessments indicate that diffusion models produce smooth, anatomically plausible boundaries while preserving global structural coherence. This eliminates the scattered errors typically associated with baseline models.

Mechanistic: Grad-CAM visualizations reveal that diffusion encoders develop bounded, organ-specific attention patterns even when in a frozen state. This thereby demonstrating that unsupervised anatomical learning takes place during the pretraining phase.

Practical: The remarkable data efficiency (achieving 95.5\% liver performance with only 10\% of labels) addresses significant challenges to clinical implementation in resource-limited environments.

Achieving 87\% of fine-tuned liver performance without utilizing labels serves as compelling evidence that the generation-as-pretext-task is effective. The denoising objective forces networks to internalize anatomical structures as a prerequisite for successful reconstruction. When these learned priors are applied to segmentation tasks, they provide substantial regularization, mitigating the risk of overfitting to misleading texture correlations and enables robust generalization across diverse anomalies. 

In addition to its direct clinical applications, this works establishes a broader principle: generative models trained on unsupervised reconstruction learn structural representations that are applicable to discriminative dense prediction tasks. As diffusion models progress, their role as unsupervised feature learners for spatial understanding tasks in medical imaging and potentially other domains deserve increasing attention from both research and clinical communities.

\section*{Acknowledgment}
The authors would like to acknowledge Manipal Institute of Technology Bengaluru, Manipal Academy of Higher Education, Manipal, India, for providing the necessary facilities and computational support to conduct this research. We express our sincere gratitude to the creators of the Beyond the Cranial Vault (BTCV) dataset \cite{gibson2018btcv} for making their data publicly accessible. Additionally, we would like to thank the developers of PyTorch, CUDA, and other open-source packages that enabled our model development and experiments.

\bibliography{references}
\bibliographystyle{IEEEtran}

\section*{Declarations}

\subsection*{A. Author Contributions Statement}
A. G. conceived the study, designed the methodology, implemented the DDPM architecture and transfer learning, performed experiments, analyzed the results, prepared the figures, and wrote the initial manuscript draft. D. G. aided in tabulating the conducted experiments results, analyzed the results, reviewed the figures and tables, and aided in the initial manuscript draft. S. B., S. A., and T. K. M. supervised the research, provided conceptual guidance and contributed to the interpretation of results, reviewed and edited the manuscript, and approved the final version.

\subsection*{B. Ethics Statement}
This article does not contain any studies with human participants or animals performed by any of the authors.

\section*{Additional Information}
\begin{itemize}
    \item \textbf{Funding:} Open access funding is provided by Manipal Academy of Higher Education.
    \item \textbf{Conflicts of Interest:} The authors declare that they have no competing financial interests or personal relationships that could have appeared to influence the work reported in this paper.
    \item \textbf{Data Availability:} The experimental evaluations were performed using the Beyond the Cranial Vault (BTCV) multi-organ abdominal CT dataset, which is publicly accessible via the official Synapse repository at \url{https://www.synapse.org/#!Synapse:syn3193805}.
    \item \textbf{ORCID iDs:} Akshat G: \url{https://orcid.org/0009-0000-9244-1513}.
\end{itemize}

\begin{IEEEbiography}[{\includegraphics[width=1in,height=1.25in,clip,keepaspectratio]{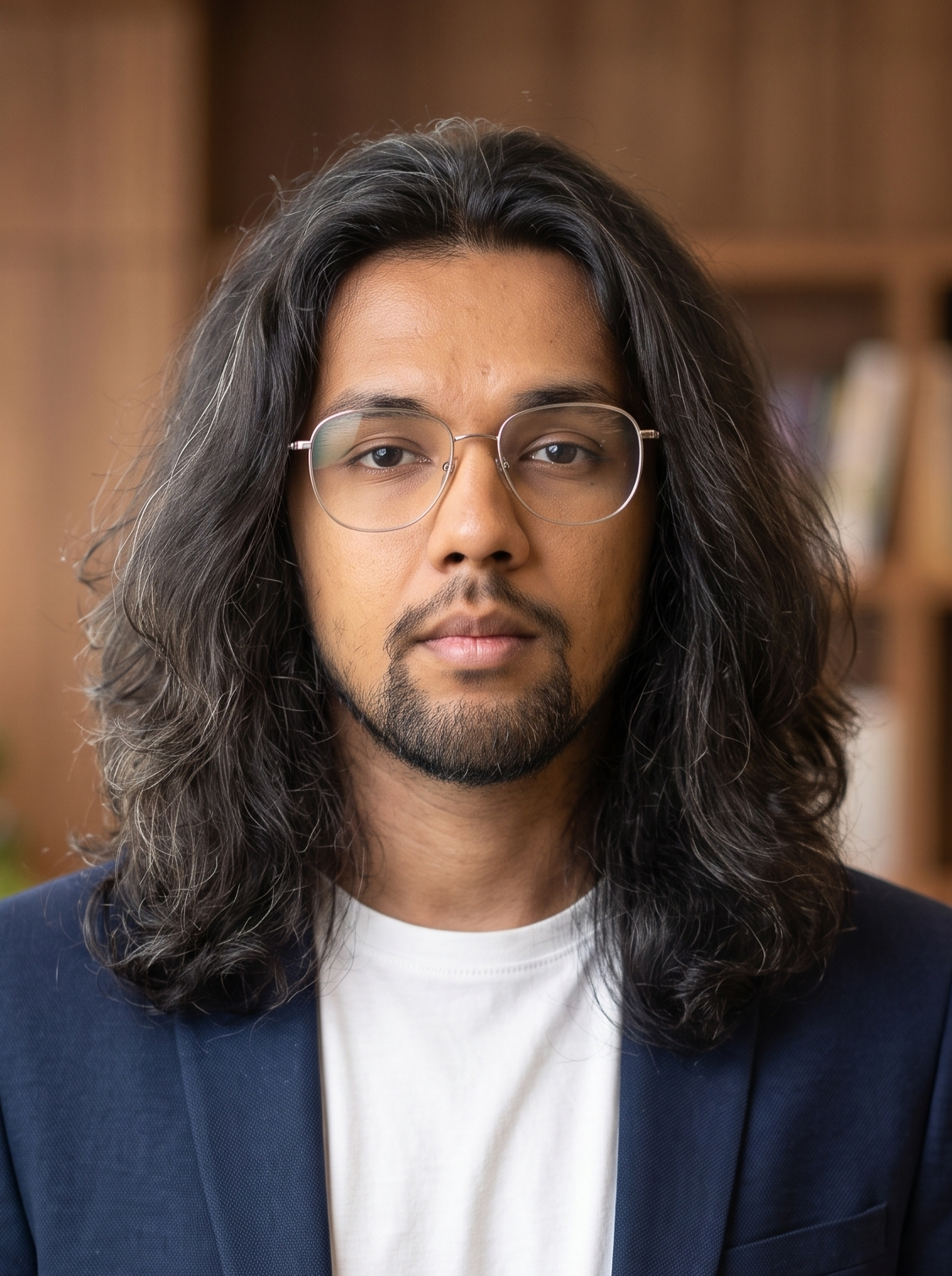}}]{Akshat G}
is currently pursuing the B.Tech Honors degree in computer science (minor in A.I. for Healthcare) with Manipal Institute of Technology Bengaluru, India. He has worked as an ML Research Intern, developing hierarchical deep learning architectures for multi label defect classification, and build end to end consumer products for his startup. His research interests include semi-supervised learning, self-supervised learning, data-centric AI, computer vision, deep learning, foundational AI and focus on developing scalable AI applications for real world challenges. 
\end{IEEEbiography}

\begin{IEEEbiography}[{\includegraphics[width=1in,height=1.25in,clip,keepaspectratio]{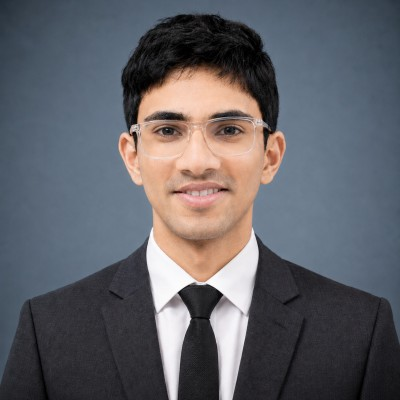}}]{Divyansh Gupta}  is currently pursuing a third-year B.Tech. degree in Computer Science and Engineering at the Manipal Institute of Technology, Bengaluru, Manipal Academy of Higher Education (MAHE), Manipal, India. He is passionate about artificial intelligence and focuses on creating scalable machine learning solutions for real-world challenges.
His areas of interest include machine learning and deep learning, particularly in the development of robust and application-oriented models.
\end{IEEEbiography}

\begin{IEEEbiography}[{\includegraphics[width=1in,height=1.25in,clip,keepaspectratio]{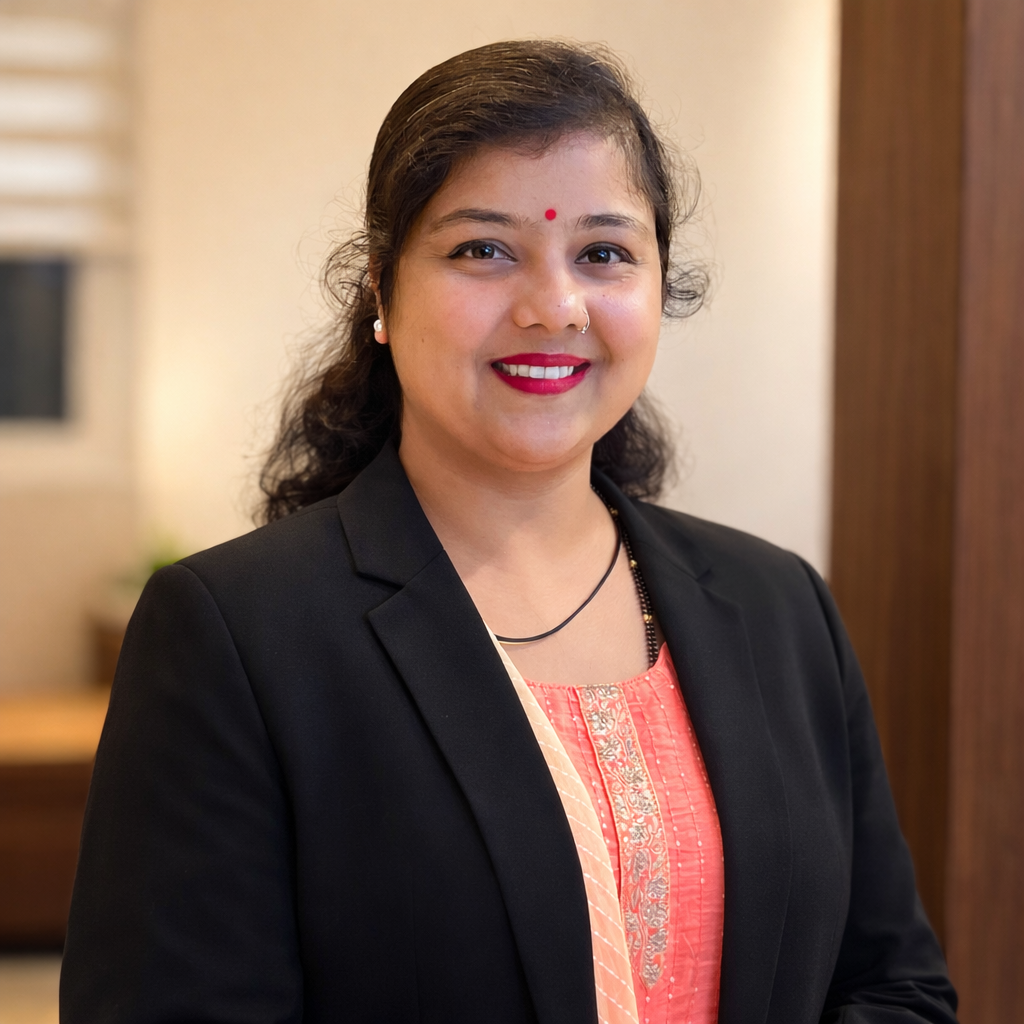}}]{Shaleen Bhatnagar} is an Assistant Professor (Senior Scale) with the School of Computer Engineering, Manipal Institute of Technology, Bengaluru, Manipal Academy of Higher Education (MAHE), Manipal, India. She has more than 14 years of experience in teaching and research.
Her research focuses on biometrics, computer vision, pattern recognition, and image processing, with recent emphasis on machine learning, deep learning, and quantum computing for real-world applications. She has authored and co-authored several research papers in these areas.
She is a Senior Member of IEEE and actively contributes to academic and professional activities within the computing research community.
\end{IEEEbiography}

\begin{IEEEbiography}[{\includegraphics[width=1in,height=1.25in,clip,keepaspectratio]{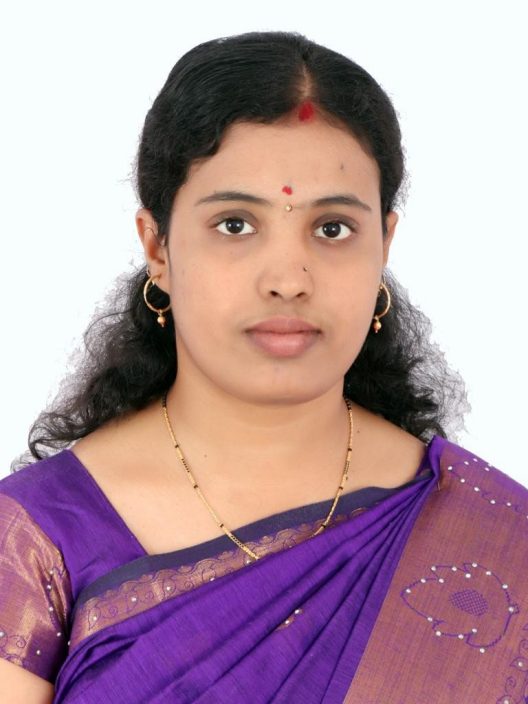}}]{Shilpa Ankalaki} received the Ph.D. degree in computer science and engineering from Visveswaraya Technological University, Belagavi, India. She is currently an Assistant Professor (Senior Scale) with the School of Computer Engineering, Manipal Institute of Technology, Manipal Academy of Higher Education, Manipal, Bengaluru. She has authored several research articles published in various international journals and conferences. Her research interests include machine learning, deep learning, data mining, artificial intelligence, and computer vision.
\end{IEEEbiography}

\begin{IEEEbiography}[{\includegraphics[width=1in,height=1.25in,clip,keepaspectratio]{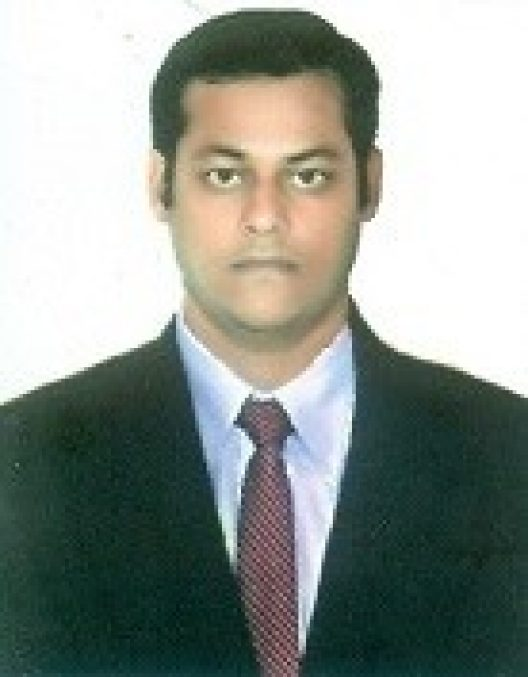}}]{Tusar Kanti Mishra}
 (Senior Member, IEEE) received the B.Tech. and M.Tech. degrees, and the Ph.D. degree from NIT Rourkela, in 2015, along with an MHRD  Fellowship. He is currently an Associate Professor with the School of Computer Engineering, Manipal Institute of Technology Bengaluru. He has more than 20 years of teaching experience. His research expertise is of ten years in the domain of artificial intelligence, computer vision, pattern recognition, along with allied application areas. He has availed the Erasmus Fellowship and worked under Prof. L. R. B. Schomaker, the Director of the ALICE Laboratory, The Netherlands, for a period of six months. He feels proud of being a student of renowned researchers, such as Dr. B. Majhi and Dr. Arun K. Pujari. He has been a reviewer for multiple reputed journals. So far, he has filed multiple patents and two copyrights with the IPR, Government of India. He has more than 35 research publications and counting. He has bagged a research funding grant of INR 30 lakhs from SERB, Government of India. He is also a member of the BoS, Sambalpur University Institute of Information Technology
\end{IEEEbiography}

\EOD

\end{document}